\documentclass[letterpaper, 10pt, journal, twoside]{IEEEtran} 

\IEEEoverridecommandlockouts

\usepackage{cite}
\usepackage{graphics}
\usepackage{epsfig}
\usepackage{mathptmx}
\usepackage{times}
\usepackage{amsmath}
\usepackage{amssymb}
\usepackage{color}
\usepackage{subcaption}
\usepackage{multirow}
\usepackage{booktabs}
\usepackage[table,xcdraw]{xcolor}
\usepackage[flushleft]{threeparttable}
\usepackage[colorlinks]{hyperref}
\usepackage{cleveref}
\usepackage{soul}
\usepackage{comment}
\usepackage{calc}

\graphicspath{{./figs/}}

\usepackage{xparse}
\DeclareMathOperator*{\argmax}{argmax}
\NewDocumentCommand\Set{m}{ \left\{#1\right\} }

\newcommand{\icratitle}{Learned Camera Gain and Exposure Control for Improved Visual Feature Detection and Matching}
\newcommand{\shorttitle}{\icratitle}

\title{\icratitle}

\author{Justin Tomasi$^{1}$, Brandon Wagstaff$^{1}$, Steven L. Waslander$^{2}$, and Jonathan Kelly$^{1}$%
\thanks{Manuscript received: October 15, 2020; Revised January 10, 2021; Accepted February 2, 2021.}
\thanks{This paper was recommended for publication by Editor Cesar Cadena Lerma upon evaluation of the Associate Editor and Reviewers' comments. 
This research was supported in part by the Canada Research Chairs program.} 
\thanks{ $^{1}$Justin Tomasi, Brandon Wagstaff, and Jonathan Kelly, a Vector Institute Faculty Affiliate, are with the Space and Terrestrial Autonomous Robotic Systems (STARS) Laboratory, University of Toronto Institute for Aerospace Studies (UTIAS), Toronto M3H 5T6, Canada. }%
\thanks{$^{2}$Steven L. Waslander is with the Toronto Robotics and AI Laboratory (TRAIL), UTIAS, Toronto M3H 5T6, Canada.}%
\thanks{\tt{<firstname>.<lastname>@robotics.utias.utoronto.ca}}%
\thanks{Digital Object Identifier (DOI): see top of this page.}
}

\begin{document}

\markboth{IEEE Robotics and Automation Letters. Preprint Version. Accepted February, 2021}
{Tomasi \MakeLowercase{\textit{et al.}}: \shorttitle}
	
\maketitle

\begin{abstract}
Successful visual navigation depends upon capturing images that contain sufficient useful information.
In this paper, we explore a data-driven approach to account for environmental lighting changes, improving the quality of images for use in visual odometry (VO) or visual simultaneous localization and mapping (SLAM).
We train a deep convolutional neural network model to predictively adjust camera gain and exposure time parameters such that consecutive images contain a maximal number of matchable features. 
The training process is fully self-supervised: our training signal is derived from an underlying VO or SLAM pipeline and, as a result, the model is optimized to perform well with that specific pipeline. 
We demonstrate through extensive real-world experiments that our network can anticipate and compensate for dramatic lighting changes (e.g., transitions into and out of road tunnels), maintaining a substantially higher number of inlier feature matches than competing camera parameter control algorithms.
\end{abstract}


\section{Introduction}

\IEEEPARstart{R}{eliable} perception is crucial for safe robot operation in dynamic environments.
While inexpensive commercial cameras have become ubiquitous due to their size, weight, and performance, camera image quality can be degraded by rapid motion and by scene lighting changes. 
In turn, poor-quality images will reduce the performance of many visual navigation algorithms and, in the worst case, may cause a navigation algorithm to fail entirely \cite{Kim2017}.

There are three general approaches to increase the robustness of visual navigation algorithms to dynamic lighting conditions \cite{Lu2010}. 
The first approach involves applying some form of post-processing after image capture in an effort to mitigate changes in illumination \cite{Clement2018, Clement2020, Gomez-Ojeda2017, Porav2018a, Park2017}.
The second approach is to utilize feature detection and matching algorithms that have some degree of invariance to brightness variations \cite{Rublee2011a, Yi2016}.
These two approaches can help to improve visual navigation when the acquired images already contain sufficient information, but cannot recover information that is lost due to overexposure or underexposure \cite{Zhang2017}.
The third approach, which we follow in this work, is to compensate for dynamic lighting during the image acquisition process by actively adjusting the relevant camera imaging parameters.
 
\begin{figure}[t!]
	\centering
	\setlength{\fboxsep}{0pt}%
	\setlength{\fboxrule}{1pt}%
	\fbox{\parbox{\columnwidth - 2pt}{%
	\includegraphics[width=0.5\columnwidth - 1pt]{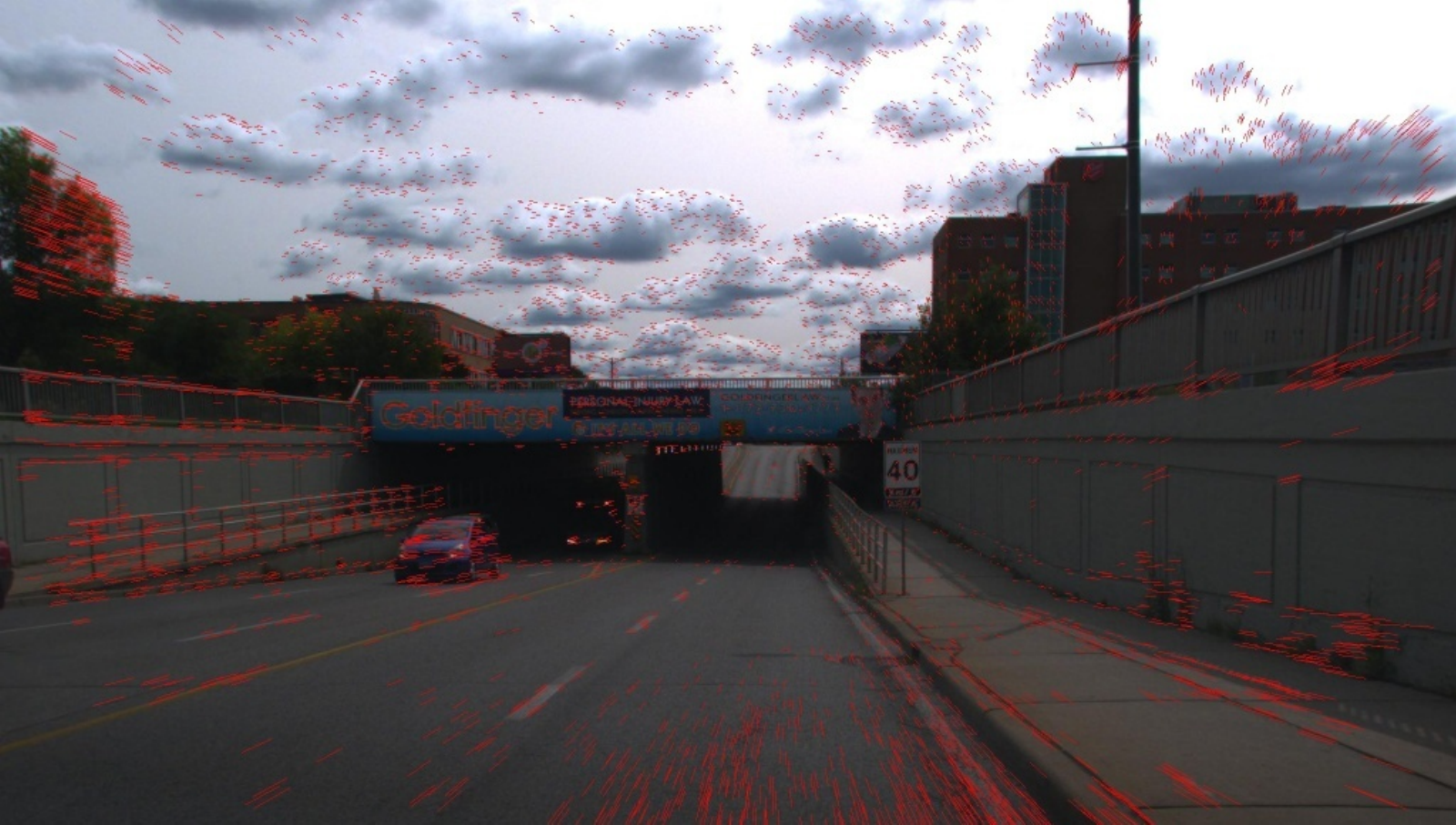}%
	\includegraphics[width=0.5\columnwidth - 1pt]{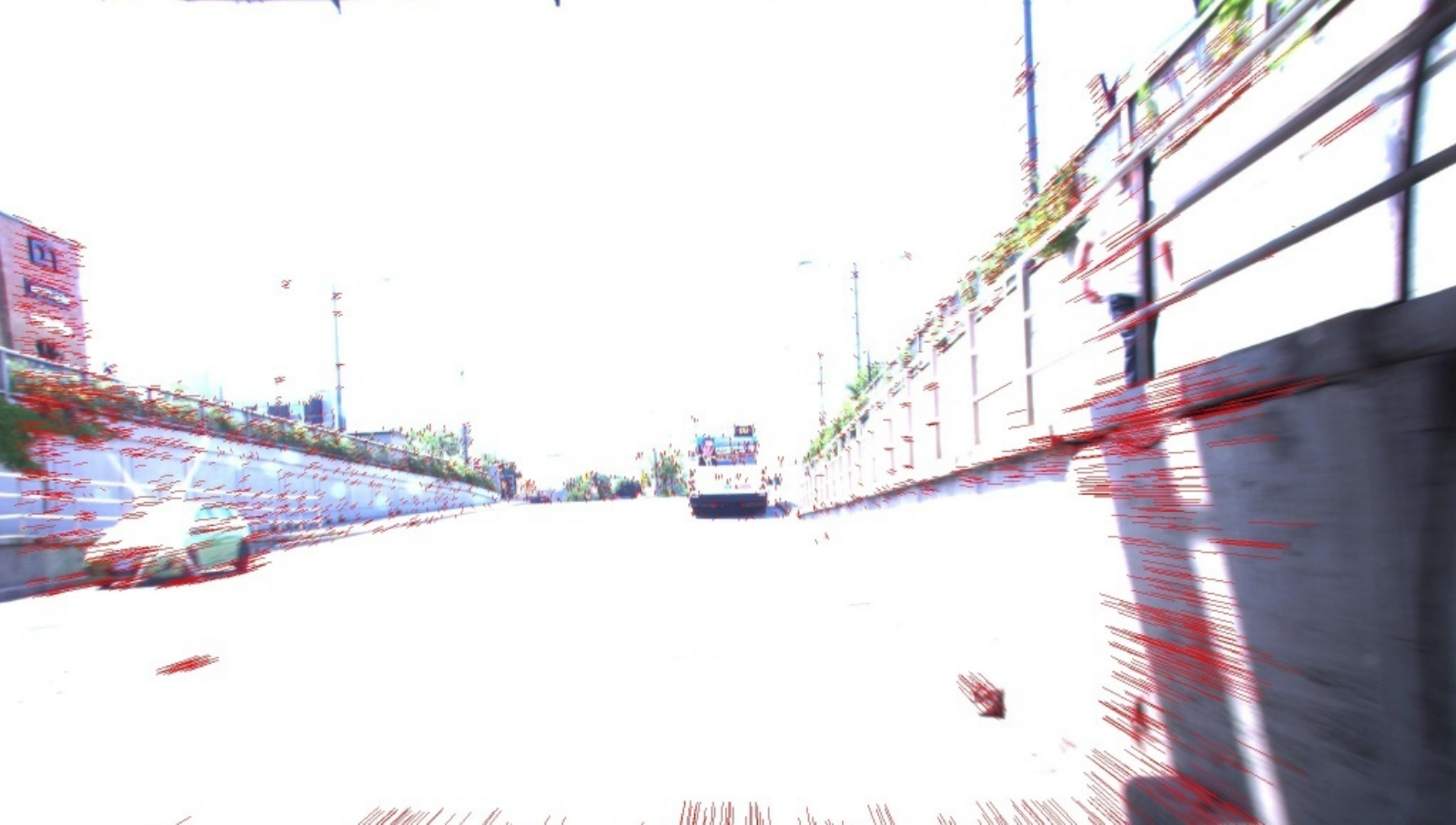}\\
	\includegraphics[width=0.5\columnwidth - 1pt]{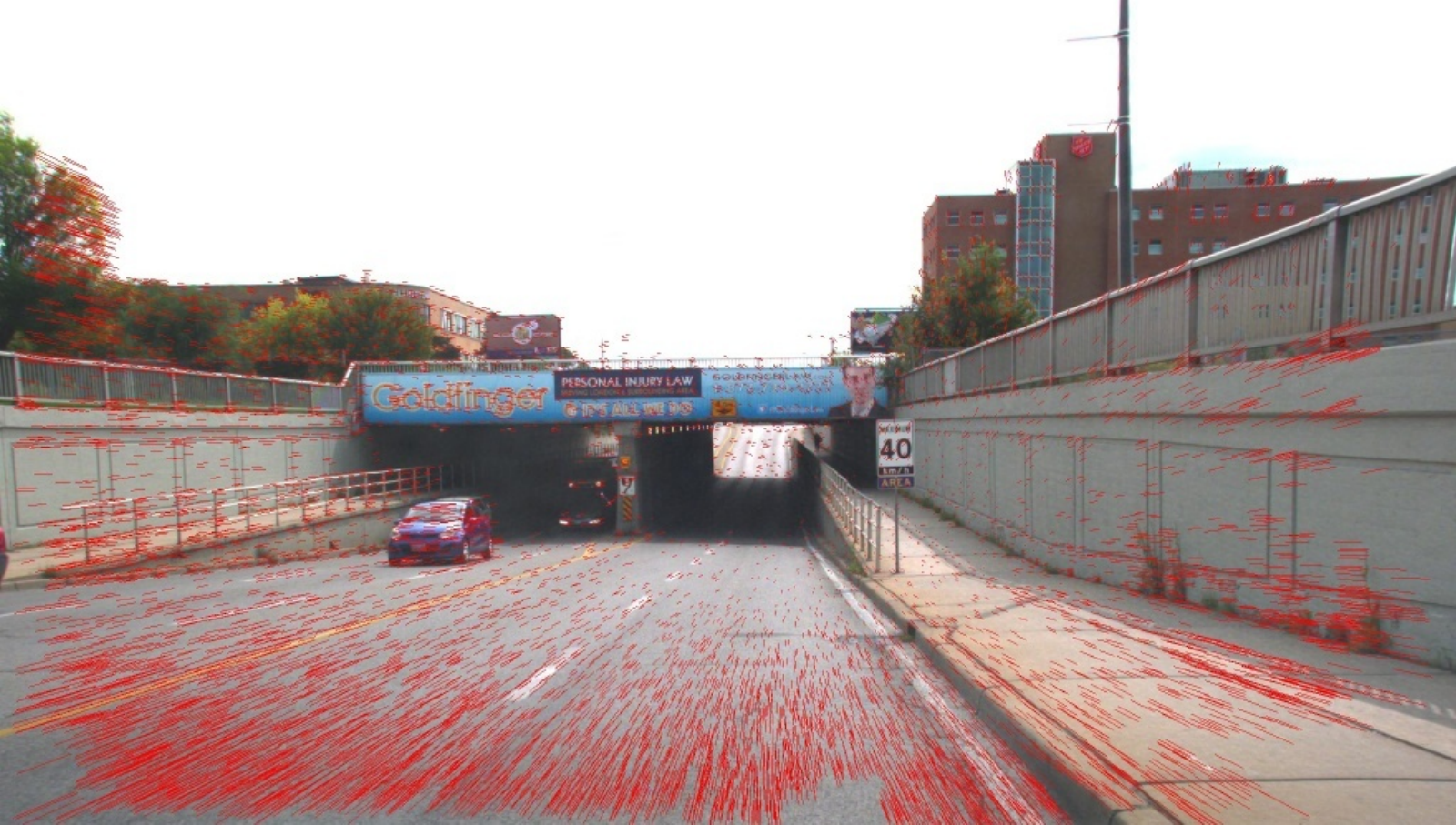}%
	\includegraphics[width=0.5\columnwidth - 1pt]{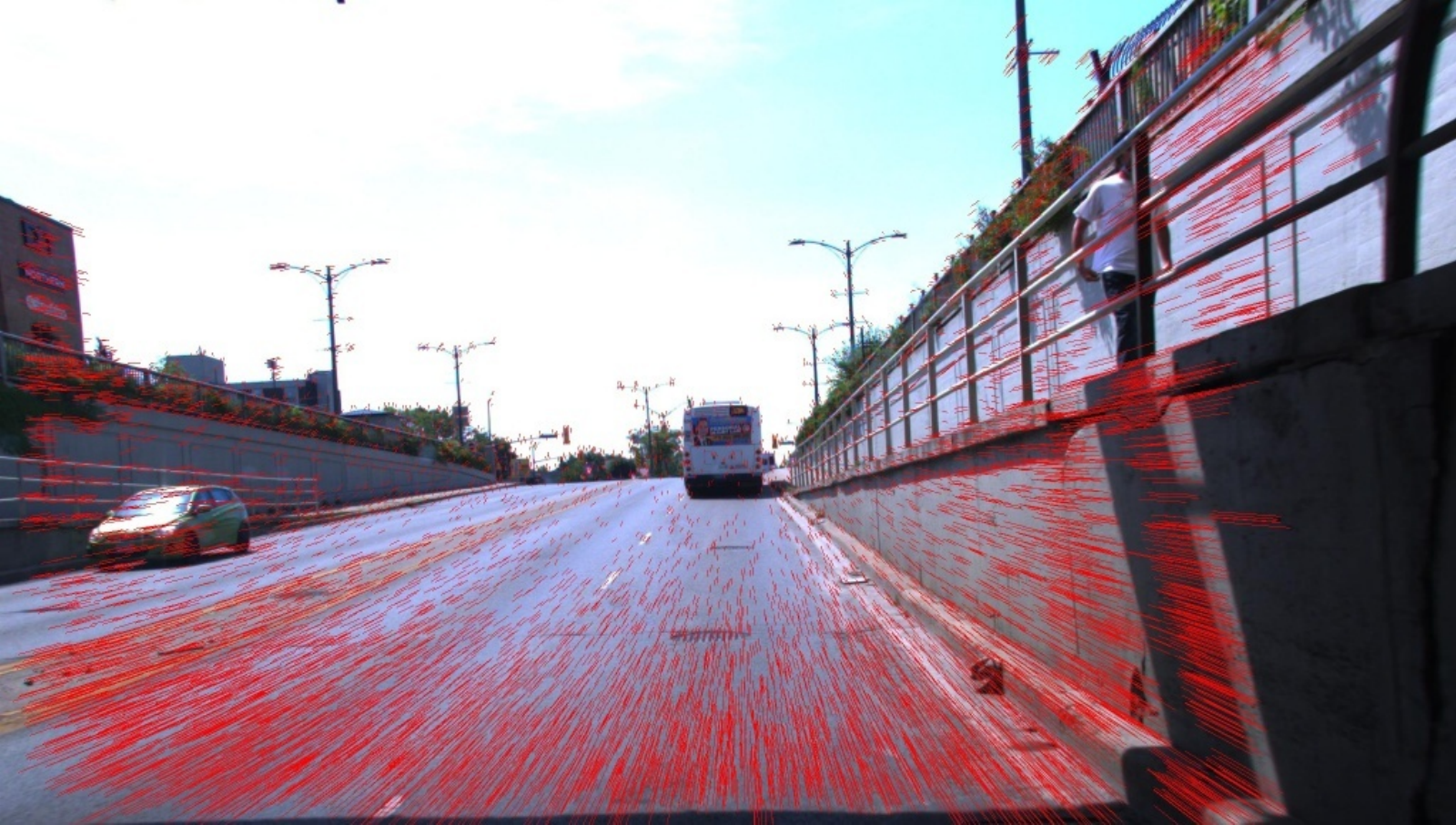}}}
	\vspace{1mm}
	\caption{Our method selects camera parameter values that yield images with a higher number of inlier feature matches (bottom row) compared to built-in automatic gain and exposure time control (top row). This behaviour is demonstrated before entering a tunnel (left column) and while exiting from a tunnel (right column). Inlier feature tracks are shown in red.}
	\label{fig1}
	\vspace{-4mm}
\end{figure}

The two camera parameters that have the greatest effect on image quality (i.e., information content) are \textit{gain} and \textit{exposure time}. 
Often, these parameters are set to fixed values for simplicity or adjusted automatically by a built-in, proprietary parameter control algorithm. 
Built-in control algorithms are usually adequate for situations in which lighting conditions are constant or change slowly. 
However, during fast lighting transitions, relying on automatic gain and exposure time control often results in poorly-exposed images due to relatively slow algorithm response times \cite{Shim2014}.

One reason for the poor performance of built-in parameter controllers and other hand-crafted algorithms is that they operate in a \emph{reactive} manner.
Adjustments are made only after a large change in overall image brightness has been recorded, which is too late to prevent the loss of valuable information (caused by overexposure or underexposure).
We posit that improved image quality under dynamic lighting conditions can be obtained through \textit{predictive} adjustments to compensate for impending lighting changes. We design such a predictive controller by training a deep neural network to adjust the camera gain and exposure time (hereafter exposure) in such a way that the quality of \textit{future} images will be improved.

Our approach is data-driven: we learn a deep convolutional neural network (CNN) model that takes as input a sequence of recent images and the corresponding camera parameter values and outputs updated parameter values that are applied before the next image is acquired.
The substantial representational capacity of deep networks allows us to capture important dependencies between scene content, lighting, and VO performance. 
For example, when trained on image data from roadways, the CNN learns to overexpose the sky in order to better expose the road region, because the sky contains limited or no useful information for navigation.
The training process is fully self-supervised and leverages the outputs of an underlying VO front end; our loss function is designed to maximize the number of inlier feature matches across consecutive images. 
Our main contributions are: 
\begin{enumerate}
	\item an algorithm for predictively adjusting camera gain and exposure parameters such that acquired images contain a greater number of sequential inlier feature matches;
	\item an approach for generating training targets in a fully self-supervised manner; and
	\item extensive real-world experimental results demonstrating that our method yields images with more inlier matches than competing parameter control algorithms.
\end{enumerate}
In particular, we demonstrate an ability to maintain successful visual tracking (and pose estimation) through road tunnel entry and exit transitions, which are challenging examples of dramatic lighting change that cause competing algorithms to consistently fail.
Although we focus on improving the performance of feature-based VO, our general approach can be altered (through an appropriate choice of loss function) to improve the quality of images for use in many different visual navigation and mapping tasks. 
 
\section{Related Work}
\label{sec:related}

Recent work in the area of camera parameter control for VO and SLAM has focused on task-agnostic, reactive adjustment to maximize some, often heuristic, measure of image `quality.' Adjustments are generally made under the assumptions that the scene content and lighting remain relatively unchanged over the adjustment period. 
In this section, we review existing methods in the literature for camera gain and exposure control to improve various image quality measures.

\subsection{Exposure Control}

Hand-crafted approaches for camera parameter control have, in many cases, focused on adjustments of exposure only \cite{Shim2014, Zhang2017, Kim2018b}. 
Exposure directly impacts image brightness and sharpness by varying the amount of light that strikes the image sensor during acquisition. 
In \cite{Shim2014}, Shim et al.\ derive an image quality metric\footnote{We note that our use of the word `metric' herein refers to a measure of image quality rather than to a distance in the mathematical sense.} that is based on the magnitude of the image gradients. 
After an image is captured, the algorithm in \cite{Shim2014} generates a series of synthetic counterparts by applying various levels of gamma correction; the optimal exposure value corresponds to the gamma correction that maximizes the gradient metric. 
Critically, however, synthetically-generated images are only able to approximate the  effects of varied exposure settings (e.g., this approach does not consider motion blur induced by longer exposures). 
Also, the method in \cite{Shim2014} is reactive---exposure adjustments are made based on the most recently acquired image only.

Similar to \cite{Shim2014}, a gradient-based image quality metric is also employed by Zhang et al.\ in \cite{Zhang2017}. 
The photometric (camera) response function is applied to model the changes in image pixel intensity that result from changes in exposure.
A gradient measure that is smoothly differentiable with respect to exposure is used in conjunction with the camera response function to determine the best exposure adjustment.
The exposure is only adjusted in the direction of the (estimated) optimal value, however, rather than directly to the optimal setting as in \cite{Shim2014}.
Additionally, photometric calibration must be carried out to determine the camera response function \cite{Debevec1997}, which may be inconvenient or impossible in many situations. 

In \cite{Kim2018b}, gradient magnitudes and the Shannon entropy of the image are combined to form a quality metric. 
Adjustments to camera exposure are made via Bayesian optimization by sampling sparsely from the parameter space. 
The sampling strategy involves acquiring images at various `test' exposure values, which may result in poor-quality intermediate images. 
Further, the method in \cite{Kim2018b} is reactive and the optimization process requires significant time. 
During rapid environmental lighting changes, prior queries of the objective surface are no longer reliable and the rate at which `optimal' images can be obtained is significantly reduced.
These factors limit the applicability of the approach for real-time scenarios, particularly in dynamic environments.

Although the adjustment of a single camera parameter reduces algorithm complexity and can, in some cases, result in improved image quality, the lack of gain control in \cite{Shim2014,Zhang2017,Kim2018b} is a significant drawback. 
For example, increases in image brightness can only be achieved through increases in exposure, which also contributes to motion blur and other detrimental effects. To acquire high-quality images in a variety of conditions, both camera gain and exposure must be controlled.

\subsection{Gain and Exposure Control}

\begin{figure*}[thpb]
	\centering
	\includegraphics[width=\textwidth]{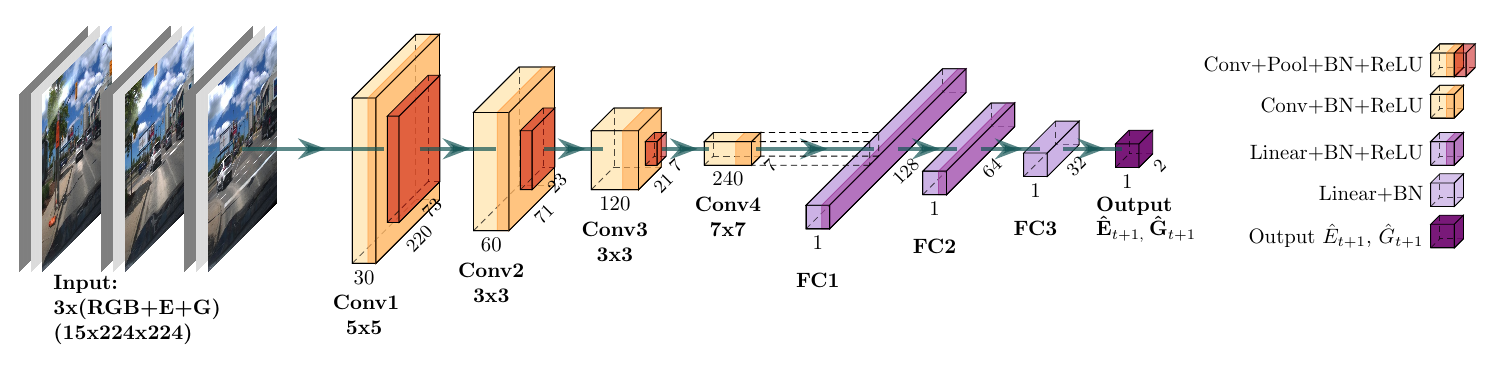}
	\vspace{-7mm}
	\caption{The structure of our predictive gain and exposure control network. The network takes as input a sequence of images $\{I_t, I_{t-1}, I_{t-2}\}$ and the corresponding gain $\{G_t, G_{t-1}, G_{t-2}\}$ and exposure $\{E_t, E_{t-1}, E_{t-2}\}$ values, and outputs the next gain $G_{t+1}$ and exposure $E_{t+1}$ settings that predictively maximize the number of inlier feature matches in future image frames.}
	\vspace{-3mm}
	\label{fig2}
\end{figure*}

The simultaneous adjustment of gain and exposure provides more flexibility to improve image quality under a wider range of conditions.
In \cite{Lu2010}, Lu et al.\ develop an algorithm to adjust gain and exposure to maximize the Shannon entropy of acquired images. 
The authors suggest that the increased entropy will result in images that contain more useful information.
The optimization of gain and exposure in \cite{Lu2010} is sampling-based: a number of images must be captured over time with varied gain and exposure settings to determine the values that maximize the image entropy.
This approach, however, is intended for use under static lighting conditions and breaks down when the lighting changes dynamically.

More recently, in \cite{Shin2019} Shin et al.\ propose to maximize an image quality metric that is a combination of the Shannon entropy, the gradient metric from \cite{Shim2014}, and a noise quantification measure.
Gain and exposure are independently adjusted through a Nelder-Mead simplex optimization  that requires sampling of real images.  
As in \cite{Lu2010}, this method is only intended for operation under static or slowly-varying lighting conditions and can fail when lighting changes occur quickly.
Unlike these existing methods \cite{Shim2014, Zhang2017, Kim2018b, Lu2010, Shin2019}, we avoid the pitfalls of reactive, sampling-based techniques through predictive, data-driven learning. At test time, our approach makes direct and immediate adjustments to both gain and exposure without the need for iteration. 

\section{Methodology}
\label{sec:method}

We train a deep CNN to predictively adjust camera gain and exposure settings in real time. Our training approach is self-supervised: we rely on an existing VO front end to extract and match features across consecutive images, and use as training targets the parameter settings that correspond to images containing a high number of features and inlier feature matches. 

\subsection{System Overview}

Our network architecture is shown in \Cref{fig2}. The network is built from a series of four convolutional blocks that incorporate max-pooling \cite{Zhou1988}, followed by three fully-connected layers. All layers, except the output, make use of batch normalization \cite{Ioffe2015} and ReLU activation functions \cite{He2016}.

The network takes as input images captured over the last three time steps, $\{I_t,\,\,  I_{t-1},\,\, I_{t-2}\}$, as well as the corresponding camera parameter settings. The images are downsampled to a lower resolution and the gain $\{G_t,\, G_{t-1},\, G_{t-2}\}$ and exposure $\{E_t,\, E_{t-1},\, E_{t-2}\}$ values at each acquisition are concatenated as additional input channels to create a 15-channel input.\footnote{Concatenation is a straightforward way to provide gain and exposure values to the network---we plan to explore more efficient network architectures in future work.}
The gain and exposure are linearly scaled to the image intensity range and assigned to every pixel in the corresponding input channel.
We make use of sequential input images to ensure that temporal information about the scene and any lighting changes is available to the network; the inclusion of the gain and exposure settings allows the network to decouple the changes due to varying parameter settings from changes due to varying external illumination. 
The network outputs the next gain $\hat{G}_{t+1}$ and exposure $\hat{E}_{t+1}$ settings to be sent to the camera.

Our network is trained with target gain and exposure values, $G_i^*$ and $E_i^*$, respectively, for the $i^\text{th}$ training sample.  
The batch loss for $N$ training samples is calculated as the weighted combination of the gain and exposure losses, where the $\epsilon$ value can be tuned to specify the importance of each parameter:
\begin{equation}
\label{eqn:loss}
\mathcal{L} = \epsilon \frac{1}{N}\sum_{i=1}^{N}|\hat{G}_{i} - G_{i}^*| + (1-\epsilon)\frac{1}{N}\sum_{i=1}^{N}|\hat{E}_{i} - E_{i}^*|,
\end{equation}

\noindent To ensure that the two terms equally contribute to the loss, we rescale the gain and exposure targets, which have an allowable range of 0--30 dB and 75~$\mu$s--30 ms, respectively, to be within the range $\left[0, 1\right]$.
At test time, we clamp the (unrestricted) network outputs to be within $\left[0, 1\right]$, and then invert the scaling as a final step.
Through empirical testing, we determined that regressing the absolute parameter values resulted in better performance than regressing a scaling or additive term applied to the current parameter values.

The targets in \Cref{eqn:loss} can be generated to suit any task-specific problem by selecting parameter values that maximize a specific objective function (or minimize a specific loss function). Herein, we develop a data labelling procedure that identifies gain and exposure settings that lead to images with a high number of identifiable features and inlier feature matches. Our motivation for this choice is discussed in the next section.

\subsection{Feature Matching as a Proxy for VO Performance}
\label{sec:vometrics}

Most techniques that seek to optimize VO performance attempt to minimize pose estimation error. However, identifying the correct camera parameters that result in the highest pose estimation accuracy is, in many situations, intractable. The real-world image acquisition process is not differentiable; we cannot employ backpropagation to update the camera parameters from captured images. Further, the majority of state-of-the-art VO pipelines are also not fully differentiable, and it is not obvious how to determine what portion of the pose estimation error should be attributed to gain and exposure adjustments.

Instead of attempting to minimize pose estimation error directly, we instead follow the approach pioneered by Clement et al. in \cite{Clement2020} and choose to maximize a proxy measure of VO performance. Our proxy measure is a combination of the number of features found in the most recently captured image and the number of inlier feature matches between sequential images (including the most recent image). This approach naturally admits a self-supervised training methodology (described in \Cref{sec:datalabelling}) in which we leverage the VO front end itself to generate our training targets. An additional advantage is that there is no requirement that the underlying pipeline be differentiable. Although existing work has shown that good VO accuracy can be achieved using small sets of features in specific instances \cite{Cvisic2018}, we note that, in general, having more features increases VO performance and robustness.\footnote{Experimental evidence for improved pose estimation accuracy with a larger number of features is provided in \cite{Tomasi2020}.}

\subsection{Dataset Collection}
\label{sec:datacollection}

We use a sampling-based data collection procedure to identify the parameter values that maximize our proxy measure.  
However, unlike existing camera parameter controllers \cite{Lu2010, Shim2014, Kim2018b, Shin2019}, no sampling is required at test time; image sampling is only carried out as part of the training process.
Further, our sampling strategy involves capturing images while moving, which ensures that we account for motion blur due to changes in exposure time. 

Our unique sampling approach leverages a dual-camera configuration, with two identical cameras mounted side-by-side (see \Cref{fig:experiment:sub2}). 
Although this technique requires additional hardware, it enables us to sample twice at each camera pose. 
While the number of samples from the parameter space at each pose is reduced compared to \cite{Lu2010} and \cite{Shim2014}, we acquire images at a relatively high frame rate (e.g., 15 Hz), which allows for a wide range of parameter values to be sampled within a short amount of time. We describe the sampling and data labelling process in more detail in \Cref{sec:datalabelling}. 

To ensure that we effectively sample values that are near, in the majority of cases, to the optimal region of the parameter space, we use an `informed' sampling approach. 
Namely, we sample around a reference set of gain and exposure values that already produce satisfactory images, rather than sampling randomly over the entire parameter space. 
When `better' quality images (i.e., those having more features or inlier feature matches) are found after perturbing the reference parameter values, the new parameter values are used for data labelling.
Otherwise, the reference parameters are used. 
To generate perturbed settings at each camera pose, the reference parameters ($G_t$ and $E_t$) from Camera 1 are independently multiplied by a random scaling factor. The perturbed parameter values are then applied to acquire an image with Camera 2. We found that applying a scaling factor of $1 \pm[0,0.5]$, where the closed interval is sampled uniformly, balances exploration of the parameter space with the possibility of sampling from sub-optimal regions.
In the case of $G_{t} = 0$ dB (which only occurs for gain), we add a small random gain value (rather than scaling). 

\begin{figure*}[thpb]
	\centering
	\includegraphics[width=1.0\textwidth]{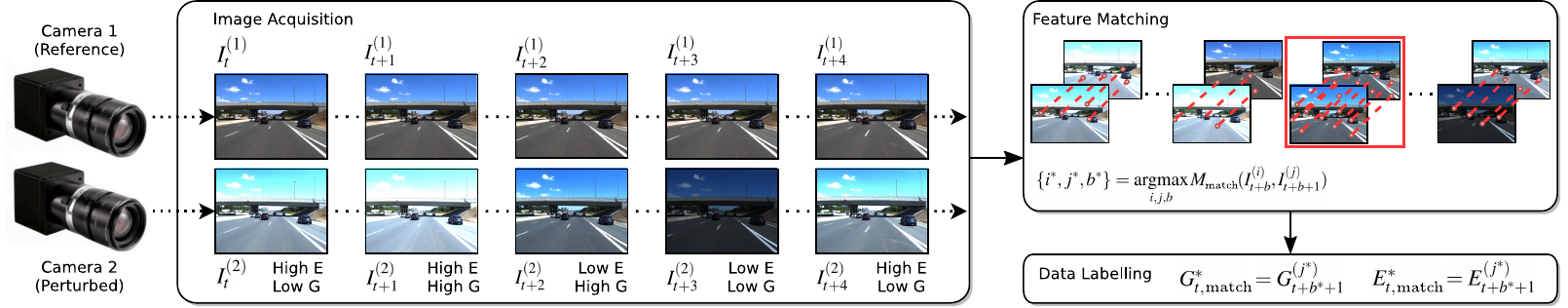}
	\caption{Our sampling-based method for generating training labels $G_{t,\text{match}}^*$ and $E_{t,\text{match}}^*$ using the $M_{\text{match}}$ method. With two cameras, four sampled image pairs (and corresponding sets of parameters) are generated at each timestep: ($I^{1}_{t}$, $I^{1}_{t+1}$), ($I^{1}_{t}$, $I^{2}_{t+1}$), ($I^{2}_{t}$, $I^{1}_{t+1}$), and ($I^{2}_{t}$, $I^{2}_{t+1}$).}
	\vspace{-2mm}
	\label{fig:datalabelling}
\end{figure*}

The sampling strategy also takes into account the direction of the perturbations. 
There are four quadrants in the parameter space that can be explored: two parameters that each can be increased or decreased.
To ensure that we effectively explore the parameter space from nearby camera poses, the four quadrants are sampled in a cycle, repeating once every four frames. 
\Cref{fig:datalabelling} depicts the image acquisition process with the reference and perturbed parameters.

An initial set of training data, collected using the built-in auto-gain and auto-exposure  controller (hereafter \textit{AG+AE}) as the reference, can be used to generate preliminary training targets through the data labelling procedure outlined in \Cref{sec:datalabelling}. 
These targets are then used to train our gain and exposure control network. 
The AG+AE reference settings, however, may not always be near the `optimal' parameter values, causing sub-optimal labels to be generated with a higher probability.
Therefore, an iterative data collection approach is used: after training our network with the initial labels, we replace the AG+AE reference with the trained network, and collect additional (perturbed) data. 
Since the trained network yields parameter settings that are closer to the `optimal' values relative to the AG+AE controller, perturbing around this new operating point is more likely to yield even higher-quality images (i.e., improved training targets).\footnote{We note the similarity of our technique to reinforcement learning and describe these similarities and differences in \Cref{sec:conclusion}.}

\subsection{Self-Supervised Data Labelling}
\label{sec:datalabelling}

In our dataset, the gain and exposure targets $G_t^*$ and $E_t^*$ are generated for an image at time step $t$ by analyzing the images in a window of future times $\{t+1,\, t+2,\, t+3,\, t+4\}$ and finding the parameters that produce the image (and image pair) that maximizes our feature count and inlier matches metrics.
We consider both the next image and images farther in the future so that there is a larger number of samples (and possible parameter values) to choose from.
A window of four poses is the minimum size required to sample from the four quadrants of the parameter space (as described in \Cref{sec:datacollection}).
Although the number of inlier feature matches in the future images may not represent the true number that would exist if we could sample repeatedly from the same pose (because the scene changes slightly with time), the windowed approach is a close approximation when the camera frame rate is reasonably high. 

Concretely, for all images within a window, we determine the number of image features, $M_{\text{feat}}(I^{(i)}_t)$,
and for all sequential image pairs within the window, we determine the number of inlier feature matches, $M_{\text{match}}(I^{(i)}_t, I^{(j)}_{t+1})$
where $i,j \in \Set{1,2}$ correspond to images from Camera 1 (reference) or Camera 2 (perturbed). We select the images that maximize these metrics, and use the corresponding parameter values as our training targets. Finally, we consider an additional metric, $M_{\text{hybrid}}$, which is a combination of the two metrics above.

\subsubsection{Generating Labels with $M_{\text{feat}}$}

The target gain and exposure values that maximize the $M_{\text{feat}}$ metric, $G^*_{t,\text{feat}}$ and $E^*_{t,\text{feat}}$ are straightforward to acquire in general. At time step $t$, each image $I^{(i)}_{t+a}$ from camera $i \in \Set{1,2}$ in the window of future time steps $a \in \Set{1,2,3,4}$ is processed using a feature detection algorithm. 
The number of features in each image is counted and we select the image $I^{(i^*)}_{t+a^*}$, where $i^*$ is the index of the camera that produced the image at time $t+a^*$ with the maximal $M_{\text{feat}}$ score:
\begin{equation}
\{i^*, a^*\} = \argmax_{i,\,a}\,M_{\text{feat}}(I^{(i)}_{t+a}).
\end{equation}
The gain and exposure values, $G^{(i^*)}_{t+a^*}$ and $E^{(i^*)}_{t+a^*}$, respectively, used to acquire $I^{(i^*)}_{t+a^*}$ are obtained and selected as the training target for time step $t$:

\begin{equation}
G_{t,\text{feat}}^* = G^{(i^*)}_{t+a^*}, \quad E_{t,\text{feat}}^* = E^{(i^*)}_{t+a^*}. 
\end{equation}

\subsubsection{Generating Labels with $M_{\text{match}}$}

Obtaining the target camera parameters for the $M_{\text{match}}$ metric, $G^*_{t,\text{match}}$ and $E^*_{t,\text{match}}$, involves performing multiple matching steps over the window of images (see \Cref{fig:datalabelling}). 
Features from both images $I^{(i)}_{t+b}$ captured by cameras $i \in \{1,2\}$ are matched with features from both images $I^{(j)}_{t+b+1}$ captured by cameras $j \in \{1,2\}$, for a particular pair of sequential time steps, $b \in \Set{0,1,2,3}$. 
We select the image pair $(I^{(i^*)}_{t+b^*}, I^{(j^*)}_{t+b^*+1})$ that results in the maximal $M_{\text{match}}$ score:
\begin{equation}
\{i^*, j^*, b^*\} = \argmax_{i,\, j,\, b}\,M_{\text{match}}(I^{(i)}_{t+b}, I^{(j)}_{t+b+1}).
\end{equation}
Note that the selected image pair may include images acquired from the same camera ($i = j$) or different cameras ($i \neq j$), depending on which combination contains more inlier feature matches.
The gain and exposure values, $G^{(j^*)}_{t+b^*+1}$ and $E^{(j^*)}_{t+b^*+1}$, used to acquire the image with camera $j^*$ at time step $t+b^*+1$ are selected as the training targets at time step $t$: 
\begin{equation}
G_{t,\text{match}}^* = G^{(j^*)}_{t+b^*+1},\quad 
E_{t,\text{match}}^* = E^{(j^*)}_{t+b^*+1}.
\end{equation}

\subsubsection{Combination of Feature Metrics}
The $M_{\text{feat}}$ metric generally yields images that are bright and well-exposed, as such images typically contain the most features. 
Consequently, this metric is well-suited for generating targets across lighting transitions. 
However, due to the sparsity of parameter sampling in our dataset, use of this metric can result in sequential targets that are quite variable. This situation occurs in particular under static lighting conditions and may lead to a low number of feature matches between consecutive frames. Conversely, the $M_{\text{match}}$ metric generally yields images that have relatively consistent gain and exposure settings across frames. Consequently, this metric is well-suited for generating training targets under static conditions. 
Generating targets with $M_{\text{match}}$, however, is less suitable during lighting transitions, as the metric favours maintaining existing feature tracks for longer durations. In turn, the changes to gain and exposure are small, and fewer new features can be found because image regions may become over- or underexposed.
We aim to balance the responsiveness of the $M_{\text{feat}}$ metric and the stability of the $M_{\text{match}}$ metric through the use of a combined metric, designated as $M_{\text{hybrid}}$. The gain and exposure values selected using the $M_{\text{hybrid}}$ metric are the weighted average of the parameter values obtained using the $M_{\text{feat}}$ and $M_{\text{match}}$ metrics.

\section{Experiments}
\label{sec:experiments}
In this section we describe our hardware platform and provide details about the environments in which we collected training data and obtained our experimental results. We then discuss the specifics of our training process and explain how we evaluated the performance of our network.

\subsection{Hardware and Configuration}
\label{subsec:hardware}

Our dual-camera setup consisted of two FLIR Blackfly S U3-31S4C machine vision cameras, each with a Fujinon 6.23 mm focal length C-mount lens, mounted side-by-side to a rigid platform on the roof of our test vehicle (\Cref{fig:experiment:sub1}) in a fronto-parallel configuration (with a baseline of 3.92 cm), as shown in \Cref{fig:experiment:sub2}.
Images were captured synchronously from both cameras.
Image capture and network processing were carried out using a Lenovo Legion Y730 laptop with an Intel i7-8750H CPU (2.20 GHz) and an NVIDIA GeForce GTX 1050 Ti GPU. With this hardware, the maximum input processing rate of the network was 640 Hz. In practice, we were limited by the rate of image acquisition ($\sim$15 Hz).

\subsection{Data Collection and Experiment Environments}

One appreciable challenge with online parameter adjustment is that the performance of the controller cannot be evaluated with previously-captured data. 
This issue arises because changes to the parameter settings affect the image acquisition process itself.
Furthermore, training data cannot be accurately simulated.
Thus, our data and results were obtained from driving in real-world conditions in all cases.

We drove our test vehicle on roads with several tunnels in the cities of London and Toronto, Ontario, Canada, under a range of outdoor illumination conditions (bright sun, low-level clouds, etc.). 
Since changes in brightness of up to 120 dB may occur during outdoor tunnel transitions \cite{Westerhoff2015}, tunnels are ideal environments for stress testing our predictive parameter controller.

\begin{figure}
	\setlength{\fboxsep}{0pt}%
	\setlength{\fboxrule}{1pt}%
	\centering
	\begin{subfigure}[]{0.45\columnwidth}
		\fbox{\includegraphics[width=\columnwidth - 2pt]{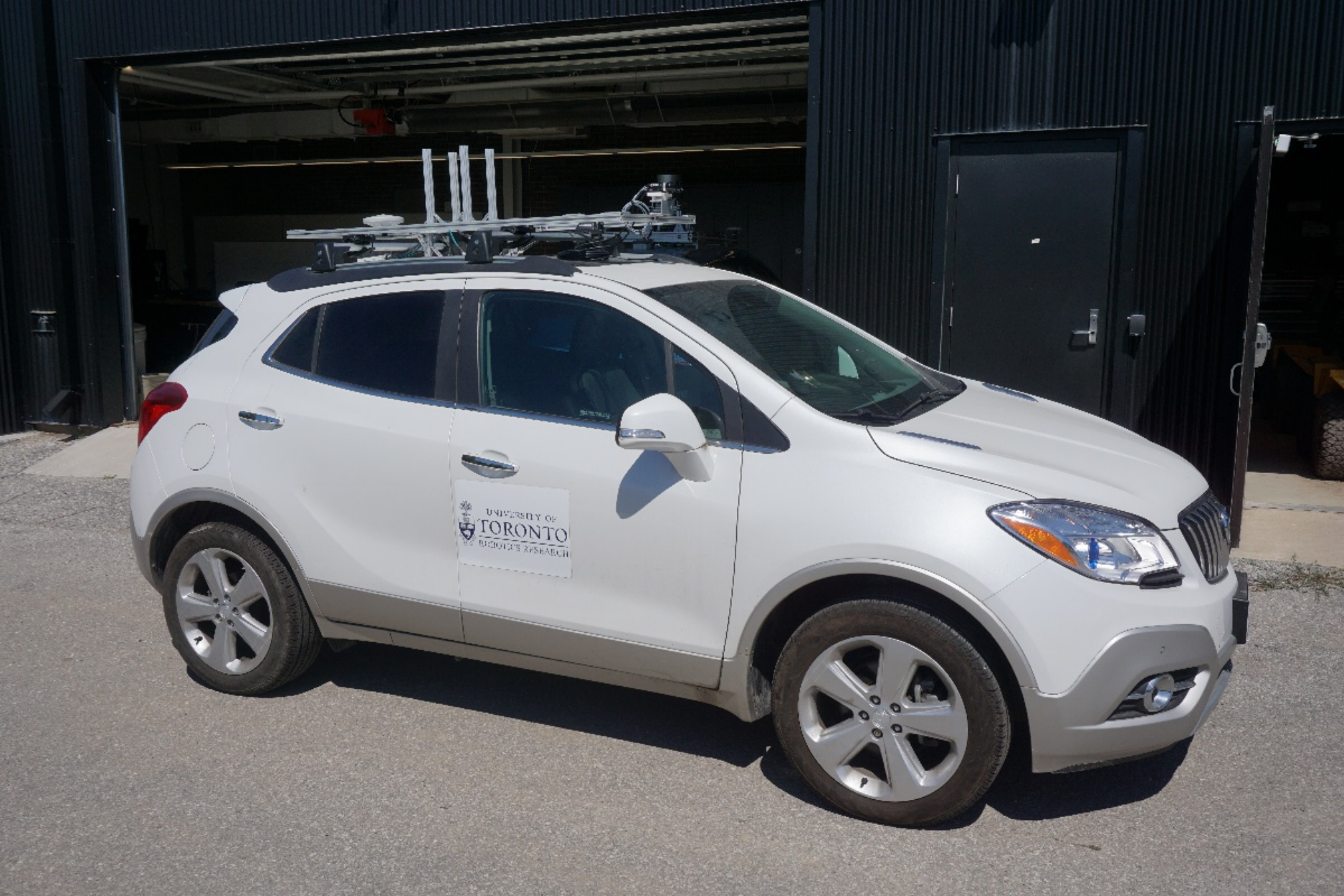}}
		\caption{Data collection vehicle.}
		\label{fig:experiment:sub1}
	\end{subfigure}
	\begin{subfigure}[]{0.45\columnwidth}
		\fbox{\includegraphics[width=\columnwidth - 2pt]{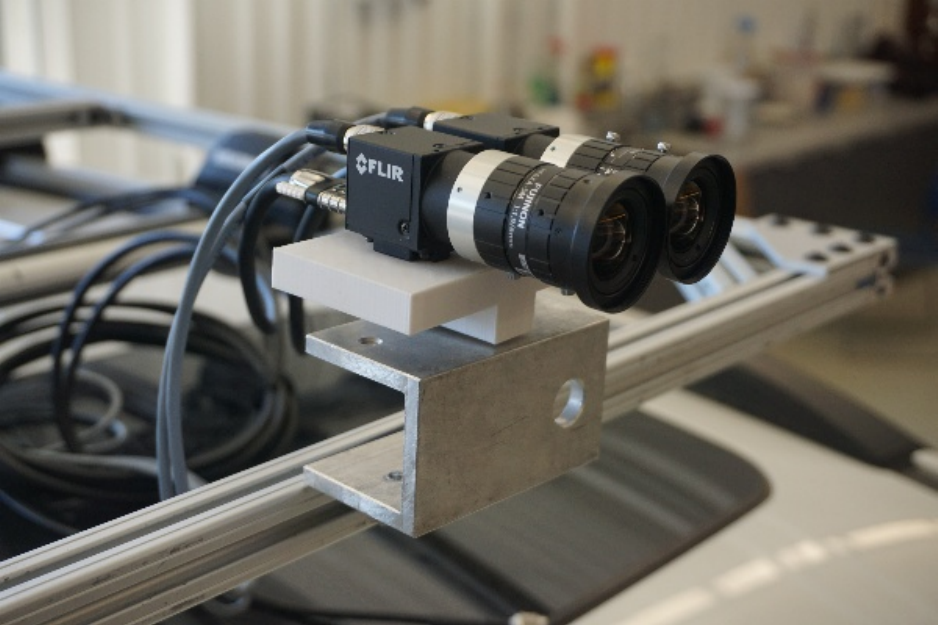}}
		\caption{Dual camera rig.}
		\label{fig:experiment:sub2}
	\end{subfigure}\\
	\vspace{2mm}
	\begin{subfigure}[]{0.45\columnwidth}
		\fbox{\includegraphics[width=\columnwidth - 2pt]{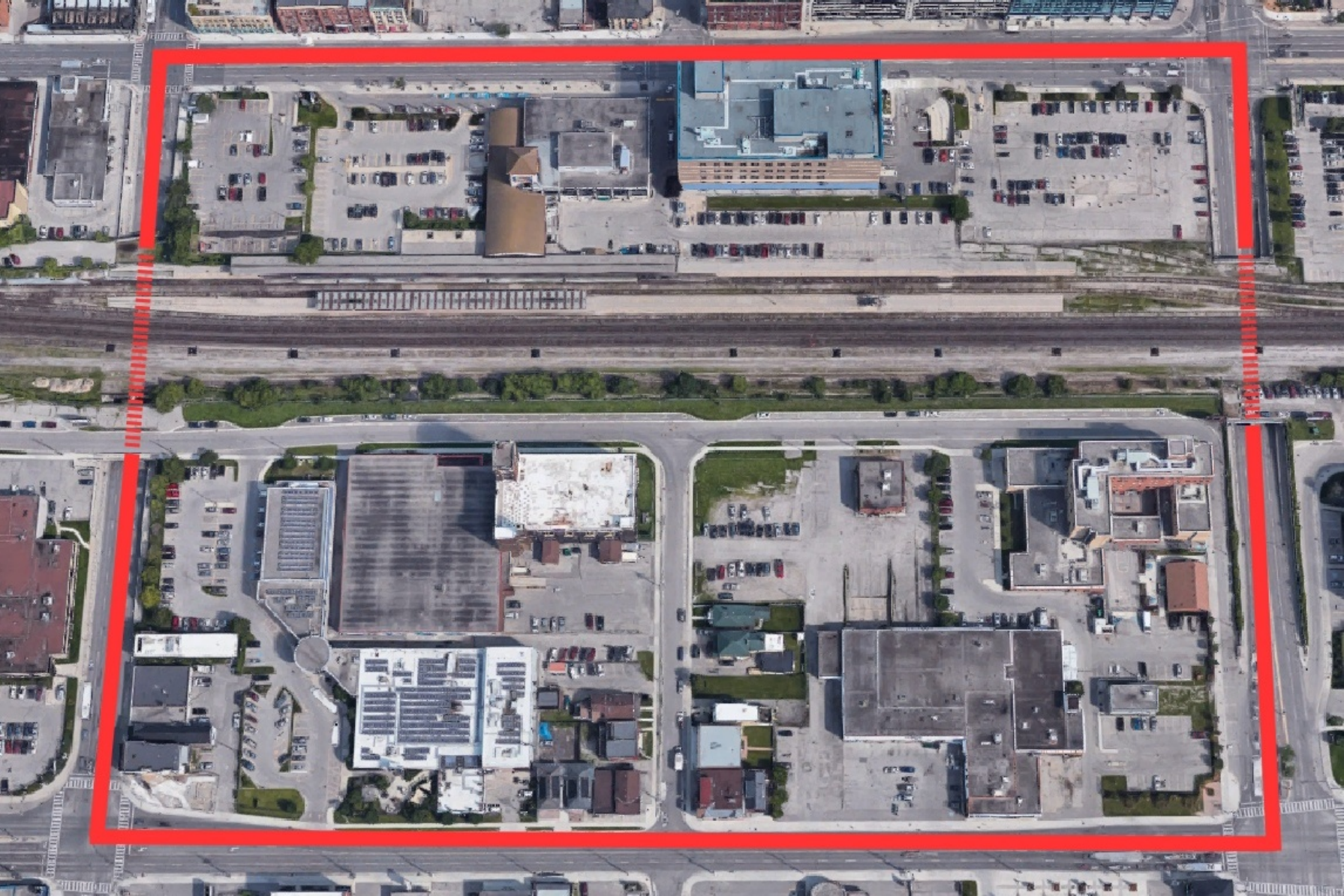}}
		\caption{London route aerial view.}
		\label{fig:experiment:sub3}
	\end{subfigure}
	\begin{subfigure}[]{0.45\columnwidth}
		\fbox{\includegraphics[width=\columnwidth - 2pt]{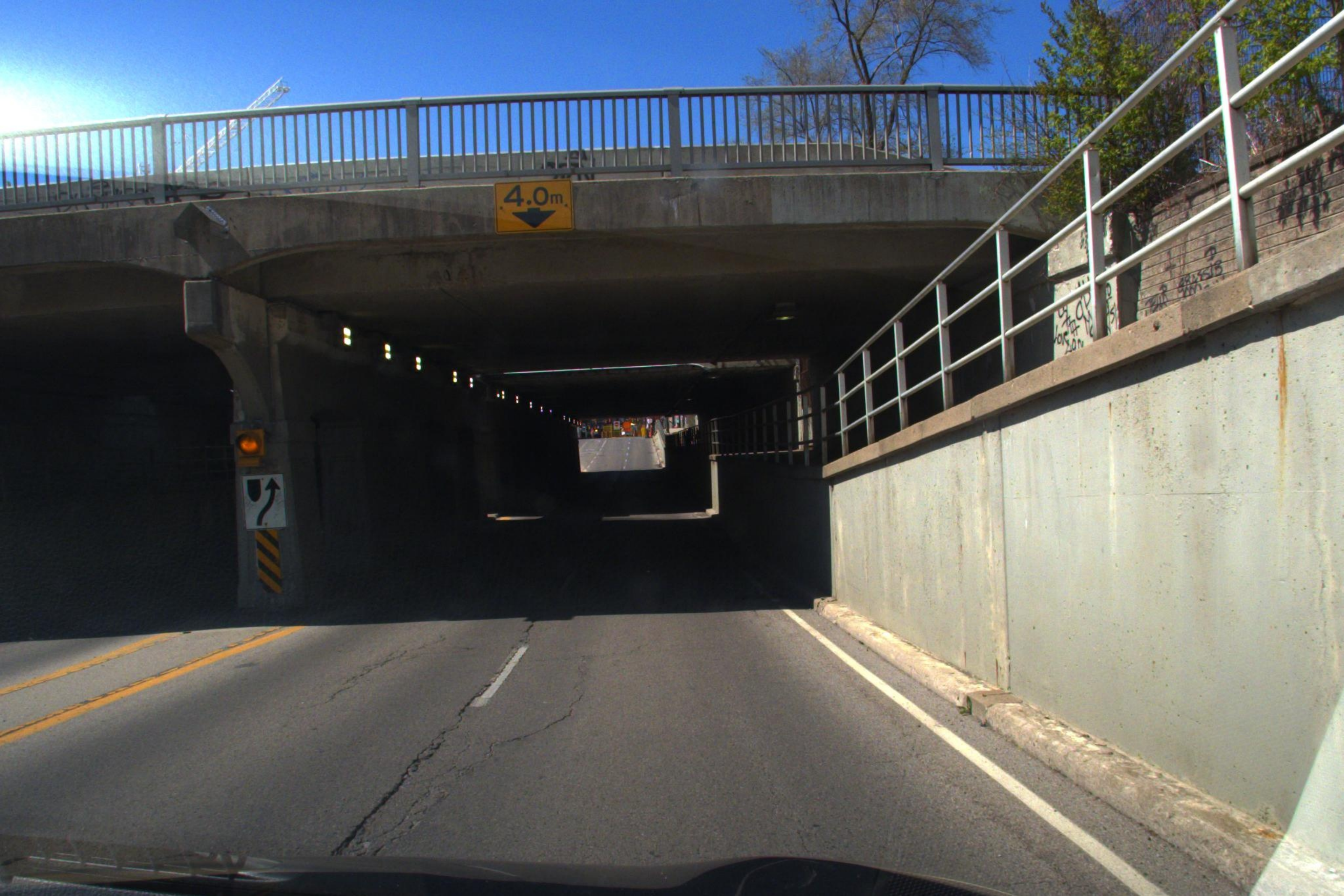}}
		\caption{London tunnel entrance.}
		\label{fig:experiment:sub4}
	\end{subfigure} 
	\caption{Photographs of the experimental setup and of the London, Ontario validation road route (which included two tunnels).}
	\label{fig:experiment}
	\vspace{-4mm}
\end{figure}

For our training dataset collection, we selected a closed route in London (see \Cref{fig:experiment:sub3}) that contains two tunnel passages (\Cref{fig:experiment:sub4}), each of roughly eighty metres in length. 
Additionally, the route contains straight sections of unobstructed road where lighting conditions are relatively constant. 
Our training data consisted of trajectories acquired along the above route as well as on a variety of other roadways in London.
After training, we experimentally validated the performance of the network on the closed route. Our held-out test environment consists of a highway exit ramp tunnel in Toronto, for which the structure, appearance, and length is significantly different from the training or validation trajectories. Overall, we collected eight validation sequences and two test sequences, for a total of 18 tunnel traverses.

\subsection{Training and Dataset Details}

Our training dataset consists of a total of fifty-five sequences containing 58,782 RGB images acquired at a 2048 $\times$ 1536 pixel resolution, collected on four different days over a period of three months.
The first set of training data includes 47,810 images and was collected using the AG+AE reference controller.
The remaining 10,972 images were collected with our network running (trained with samples from the first set of data) to control the reference camera.
We found that two training iterations was sufficient and that further iterations produced diminishing returns.

Training and validation samples were generated from our data in the manner described in \Cref{sec:datalabelling} using the $M_{\text{hybrid}}$ metric, with an equal weighting between the maximized $M_{\text{feat}}$ and $M_{\text{match}}$ metrics.
Training images (seen in \Cref{fig2}) were downsampled to a resolution of 224 $\times$ 224 pixels. Each training sample required three sequential input images: to generate the sample, we selected one of the two available images (i.e., from either camera) at each of the three time steps (yielding eight possible sequences). We maximized our data efficiency by using all eight combinations for training, resulting in $N = 232,934$ total training samples; this also ensured that the network was trained with images acquired with a wide range of gain and exposure combinations.
Our network was trained for 200 epochs using the Adam optimizer \cite{Kingma2015} with a batch size of 64 and learning rate of $1 \times 10^{-4}$. 
We also made use of dropout \cite{Srivastava2014} ($p=0.4$) in each layer to improve generalization.
We selected $\epsilon = 0.5$ in the loss function to equally weigh the gain and exposure losses, under the assumption that gain and exposure are roughly equally important in determining image quality.
The training hyperparameters were determined empirically through a random-search tuning strategy on a held-out set of training data. 

\begin{figure*}[thpb]
	\setlength{\fboxsep}{0pt}%
	\setlength{\fboxrule}{1pt}%
	\fbox{\includegraphics[width=\textwidth - 2pt]{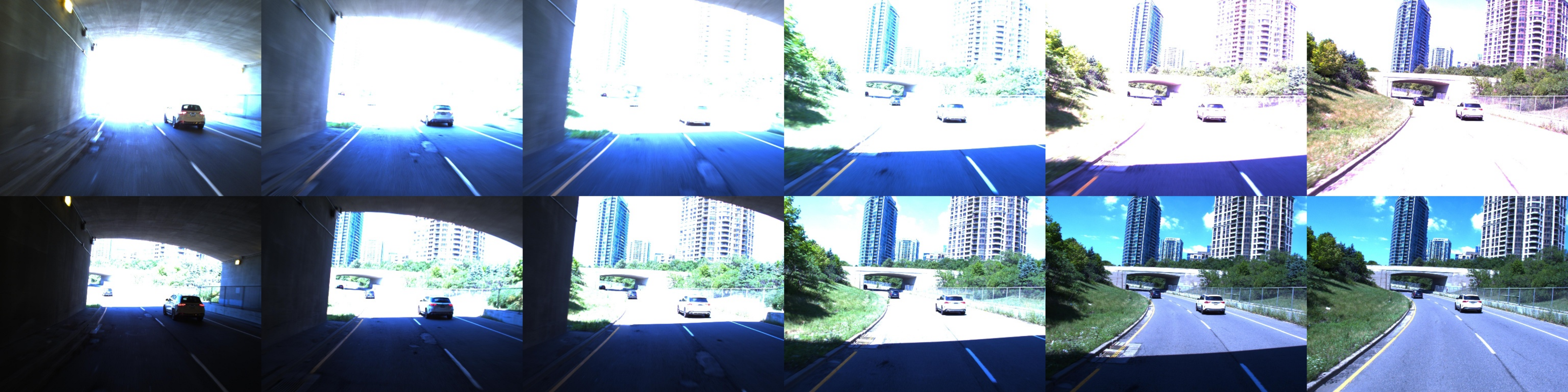}}
	\centering
	\caption{A sequence of images acquired at test time during a transition out of a tunnel along the Toronto route. Our approach (bottom row) compensates for the drastic change in lighting and maintains a higher number of inlier feature matches across all images during the transition when compared with AG+AE (top row).}
	\vspace{-4mm}
	\label{fig:network_sequence}
\end{figure*}

\begin{figure}[b!]
	\vspace{-4mm}
	\centering
	\includegraphics[trim=24pt 18pt 0pt 0,clip,width=\columnwidth]{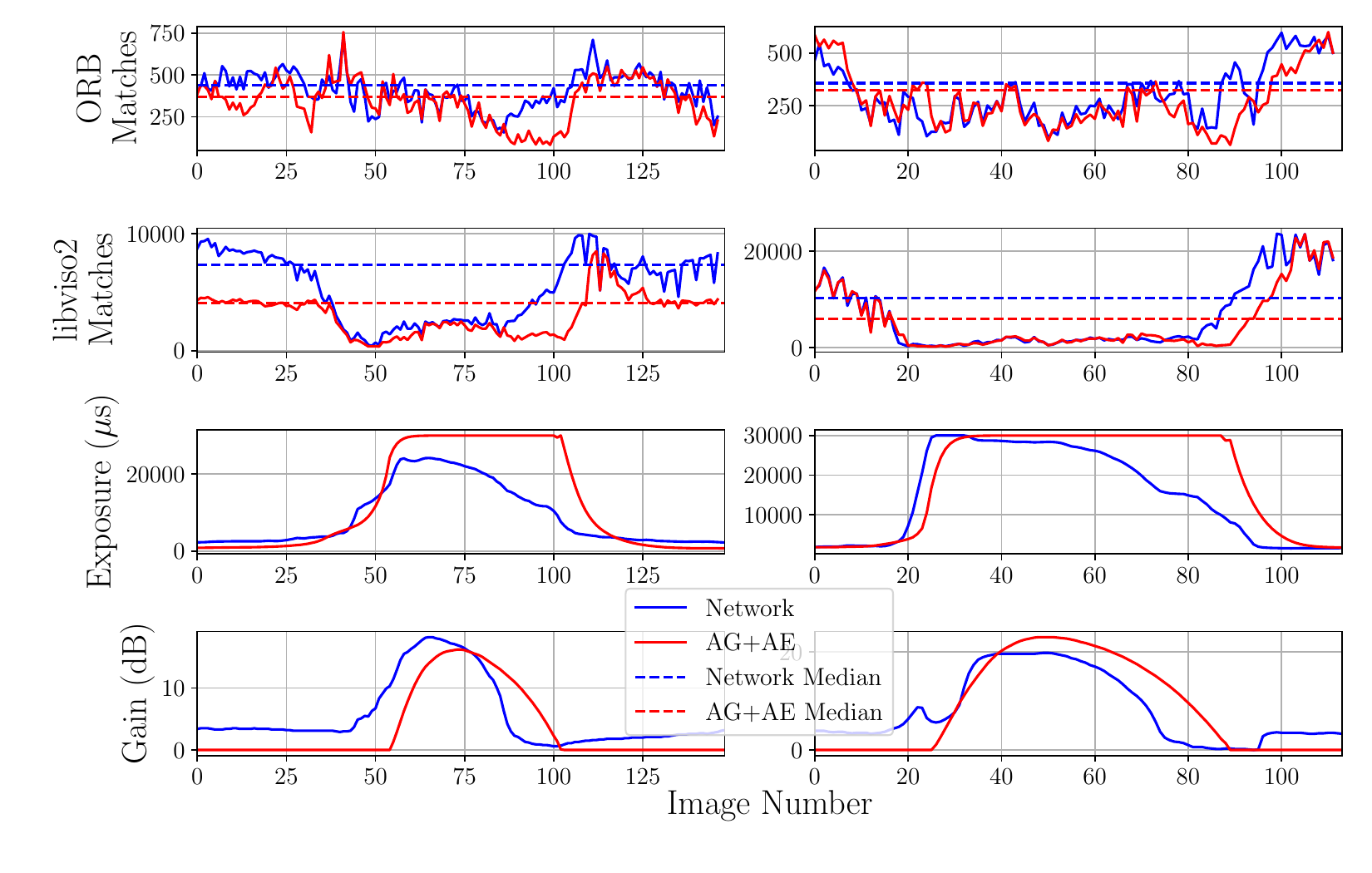}
	\caption{ORB and \texttt{libviso2} inlier feature match statistics plotted over time for a London validation sequence (left) and a Toronto test sequence (right) tunnel transition. The corresponding per-frame gain and exposure settings are also shown. Our method preemptively adjusts gain before entering and exiting the tunnel and maintains a lower exposure setting, leading to a reduction in blur and more inlier feature matches compared with AG+AE.}
	\label{fig:matchplot}
\end{figure}

\subsection{Network Evaluation}

To evaluate the performance of our network-based controller, we conducted an extensive set of real-world experiments, where we measured the number of inlier feature matches in images acquired using our method. 
We compared the performance of our controller to both built-in AG+AE and the method of Shin et al.\ \cite{Shin2019}, and also investigated combining these controllers with a type of illumination-invariant image transformation. 

Our validation experiments involved capturing full traversals of the selected route in London (\Cref{fig:experiment:sub3}), starting and ending at roughly the same pose.
We subdivided each sequence into two categories: `dynamic', which corresponds to tunnel regions, and `static', and compared the performance of the controllers on each subsequence.
For every route traversal, we selected two dynamic sections and two static sections. 
Our test-time experiment involved acquiring images while driving into and out of a highway tunnel in Toronto.
In all cases, the captured images were processed using the OpenCV ORB \cite{Rublee2011a} and \texttt{libviso2} \cite{Ji2016} feature matching algorithms.
Additionally, we repeated the matching experiments after transforming all images using the ``SumLog'' image transformation outlined in \cite{Clement2020}. We tuned the transformation parameters using a subset of our training dataset by selecting the parameter values that yielded the maximum number of feature matches between sequential image pairs.

To evaluate the success of our method, we recorded the median number of inlier feature matches (median NFM) and the minimum number of inlier feature matches (minimum NFM) on a per-image basis over sections of the recorded trajectories. 
The minimum NFM is an important performance measure because VO may fail in cases where the NFM is too low. 
A low minimum NFM typically occurs when a series of consecutive frames are over- or underexposed.
The median NFM provides an indication of the expected `average' performance of VO over the trajectory.

Finally, we sought to determine if our method improves the overall robustness of VO.
To do so, we processed the images from the validation and test sequences using \texttt{ORB-SLAM2} \cite{mur2015}, which fails when images do not contain sufficient numbers of matchable features (e.g., in our case, images acquired near tunnel entrances and exits).
We recorded the number of sequences for which \texttt{ORB-SLAM2} was able to maintain successful tracking throughout.
We expect that, within dynamic lighting regions, images acquired using our network-based controller will yield more reliable \texttt{ORB-SLAM2} outputs compared to images acquired using the built-in AG+AE controller.

\section{Results}

We summarize our feature matching results for the validation sequences in \Cref{tab1} and the test sequences in \Cref{tab2}. 
Under static lighting conditions, all of the parameter controllers were able to obtain large median and minimum NFM scores, however, our network generally produced images containing more matchable features.
The advantages of our approach are most noticeable in the dynamic experiments. 
Here, our network obtained significantly higher median and minimum NFM scores, especially compared with the Shin algorithm \cite{Shin2019}, which failed to find suitable parameter settings during the fast tunnel transitions.

\begin{table}[t]
	\centering
	\caption{A comparison of our network with AG+AE (six traversals) and the Shin method \cite{Shin2019} (two traversals) over the London validation sequences. We recorded the median NFM and minimum NFM, averaged across all sequences, with an average of $\sim$250 images in each dynamic sequence and $\sim$225 images in each static sequence.}
	\vspace{0mm}
	\label{tab1}
	\begin{threeparttable}
	\begin{tabular}{@{}llllll@{}}
			\toprule
			Lighting & Method & \multicolumn{2}{l}{Median NFM} & \multicolumn{2}{l}{Minimum NFM} \\ \midrule
			& 	                          & \texttt{ORB} & \texttt{libviso2} & \texttt{ORB} & \texttt{libviso2}\\ \cmidrule{3-6}
			\multirow{7}{*}{\textbf{Static}} 
			& AG+AE                       & 593          & 8301              & 366          & 5840 \\
			& Ours                        & \textbf{600} & \textbf{9591}     & \textbf{401} & \textbf{6636}\\ \cmidrule{3-6}
			& Shin \cite{Shin2019}        & 414          & 5295              & 183          & 3555\\
			& Ours                        & \textbf{439} & \textbf{5666}     & \textbf{200} & \textbf{4032} \\ \cmidrule{3-6}
			& AG+AE (SL)              & 436          & \textbf{6005}     & 171          & \textbf{4242}\\
			& Ours (SL)               & \textbf{445} & 5862              & \textbf{175} & 4174 \\ \midrule
			\multirow{7}{*}{\textbf{Dynamic}} 
			& AG+AE                       & 393          & 5350              & 54           & 301 \\
			& Ours                        & \textbf{395} & \textbf{7380}     & \textbf{98}  & \textbf{421}\\ \cmidrule{3-6}
			& Shin \cite{Shin2019}        & 131          & 614               & 0            & 0\\
			& Ours                        & \textbf{202} & \textbf{3530}     & \textbf{37} & \textbf{127} \\ \cmidrule{3-6} 
			& AG+AE (SL)              & \textbf{264} & 3347              & 10          & 235\\
			& Ours (SL)               & \textbf{264} & \textbf{3680}     & \textbf{40} & \textbf{385} \\ \bottomrule
		\end{tabular}
	\end{threeparttable}
\end{table}
	
\begin{table}[t]
	\centering
	\caption{A comparison of inlier feature match statistics for our network compared with AG+AE for the Toronto test sequences.}
	\vspace{0mm}
	\label{tab2}
	\begin{threeparttable}
	\begin{tabular}{@{}llllll@{}}
			\toprule
			Lighting & Method & \multicolumn{2}{l}{Median NFM} & \multicolumn{2}{l}{Minimum NFM} \\ \midrule
			& 	                          & \texttt{ORB} & \texttt{libviso2} & \texttt{ORB} & \texttt{libviso2}\\ \cmidrule{3-6}
			\multirow{4}{*}{\textbf{Dynamic}} 
			& AG+AE                       & 323          & 6035              & 63          & 245 \\
			& Ours                        & \textbf{357} & \textbf{10229}    & \textbf{96} & \textbf{305}\\ \cmidrule{3-6}  
			& AG+AE (SL)              & 198          & 2684              & 17          & 154\\
			& Ours (SL)               & \textbf{279} & \textbf{4927}     & \textbf{38} & \textbf{290} \\ \bottomrule
		\end{tabular}
	\end{threeparttable}
	\vspace{-4mm}
	\end{table}

In the SumLog transformation (SL) image experiments, our method yielded higher median and minimum inlier feature matches under dynamic lighting conditions compared with AG+AE (SL), despite these scores being lower than those for the untransformed images. We attribute this in part to the SL transform being suited for matching across extreme appearance changes, rather than matching across sequential images that are already relatively similar. Our results show that existing post-processing techniques cannot recover information that is lost due to image saturation. Instead, a method such as ours that ensures appropriate parameter settings are used during image capture is vital for acquiring high-quality, matchable images.

\begin{table}[t]
	\centering
	\caption{The number of experiments in which \texttt{ORB-SLAM2} VO successfully maintained tracking.}
	\label{tab3}
	\begin{threeparttable}
	\begin{tabular}{@{}lcc@{}}
			\toprule
			 & \multicolumn{2}{c}{VO Tracking Successes}\\
			\cmidrule{2-3}
			Method & Validation Trials (London) & Test Trials (Toronto)\\ \midrule
			AG+AE & 0/6 & 0/2\\
			Ours & \textbf{6/6}& \textbf{2/2} \\ \bottomrule
		\end{tabular}
	\end{threeparttable}
	\vspace{-4mm}
	\end{table}

Examples of the operation of our controller during both a validation sequence and a test sequence tunnel transition are shown in \Cref{fig:matchplot}.
During the transition into the tunnel, beginning at frame 40 approximately (left plots) and at frame 20 approximately (right plots), our network preemptively increased the gain value and to a lesser extent, the exposure, relative to AG+AE. 
During the transition out of the tunnel, around frames 90--110 (left plots) and frames 80--100 (right plots), our network preemptively reduced the gain and exposure relative to AG+AE, resulting in images containing less motion blur and more visible details outside of the tunnel. 
Consequently, there was a dramatic increase in both ORB and \texttt{libviso2} inlier matches in this region (see \Cref{fig1} lower row).
Examples of images acquired during a Toronto test sequence tunnel exit are shown in \Cref{fig:network_sequence}.

Finally, in \Cref{tab3}, we show that \texttt{ORB-SLAM2} was able to successfully track across all validation and test sequences when our network was employed to control the camera.
Conversely, \texttt{ORB-SLAM2} consistently failed when the built-in AG+AE controller was used instead. Here, a failure was identified if, at any point in the sequence, \texttt{ORB-SLAM2} reported that too few features matches were available to compute a pose change estimate. Notably, our analysis revealed that, in all cases, tracking failed because the AG+AE images were overexposed during transitions out of tunnels.

\section{Conclusions and Future Work}
\label{sec:conclusion}

The predictive adjustment of camera gain and exposure settings can improve the quality of acquired images for use in visual navigation. 
We demonstrated that our CNN, trained in a self-supervised manner with targets generated from a feature-based VO front end, selects camera parameters that result in images containing significantly more features and sequential inlier feature matches compared with reactive algorithms, under static and dynamic lighting conditions.
Our network can predict changes in lighting due to an approaching tunnel entrance or exit, for example, and compensate for these changes by adjusting gain and exposure preemptively.
We verified that the increased number of inlier matches due to preemptive adjustment improves the robustness of feature-based VO.
Results in the literature also indicate that adjustments which improve feature matching are likely to benefit other vision tasks as well\mbox{\cite{Shim2014,Shin2019}}.
  
Although our predictive controller works well under static and dynamic conditions, we expect that our results could be further improved if a more sophisticated sampling strategy were employed, or better yet, if more images could be obtained from the same (or nearby) camera poses.
We relied on the number of features and inlier feature matches as a proxy for feature-based VO performance, but our network could also be trained with a different loss function.
For example, a dense photometric loss would improve the quality of images for use in direct VO.
Alternatively, improvements could possibly be obtained by framing camera parameter control as a reinforcement learning (RL) problem.
The iterative component of our technique is similar in spirit to RL, in the sense that we are learning a policy and then iteratively updating the policy to increase the `reward' (the number of features and feature matches). 
In practice, we developed our unique method because it was infeasible to train our network with standard RL algorithms (due to the large state space of images and to challenges involving simulation).
However, these challenges may be overcome; we leave the implementation of an RL-based framework as future work.

{\footnotesize
\bibliographystyle{IEEEtran}
\bibliography{abbrevs,refs}
}
\end{document}